\documentclass{article}
\usepackage{fancyhdr}
\usepackage{report}

\usepackage{soul}
\usepackage[utf8]{inputenc} % allow utf-8 input
\usepackage[T1]{fontenc}    % use 8-bit T1 fonts
\usepackage{hyperref}       % hyperlinks
\usepackage{url}            % simple URL typesetting
\usepackage{booktabs}       % professional-quality tables
\usepackage{longtable}      % multi-page tables
\usepackage{tabularx}       % auto-width tables
\usepackage{amsfonts}       % blackboard math symbols
\usepackage{fancyhdr}       % custom headers and footers

\newcommand{\ours}{CLI-Universe\xspace}

\usepackage{nicefrac}       % compact symbols for 1/2, etc.
\usepackage{microtype}      % microtypography
\usepackage{graphicx}
\usepackage{pifont}
\usepackage{multirow}
\usepackage{CJKutf8}
\definecolor{darkmagenta}{rgb}{0.56, 0.0, 1.0}
\definecolor{softyellow}{rgb}{1.0, 0.92, 0.3} % richer, warmer yellow
\definecolor{LightAquamarine}{rgb}{0.75, 1.0, 0.8} % soft aqua green
\definecolor{FireBrick}{RGB}{178,34,34}
\definecolor{MediumPurple}{RGB}{147,112,219}

\definecolor{uclablue}{rgb}{0.15, 0.45, 0.68}
\hypersetup{
    breaklinks,
    colorlinks=true,
    citecolor={darkmagenta},
    linkcolor={uclablue},
    urlcolor={uclablue}
}
\usepackage{wrapfig}
\usepackage{float}
\usepackage{subcaption}
\usepackage{placeins}
\usepackage{lipsum} % 用于生成示例文字
\usepackage{tcolorbox}
\usepackage{amsmath}
\usepackage{amssymb}
\usepackage{fontawesome}
\usepackage{xspace}
\usepackage{enumitem}
\usepackage{multirow} 
\tcbuselibrary{breakable}
\usepackage{enumitem}
\usepackage{colortbl}
\usepackage{fancyhdr}

\usepackage{tcolorbox}
\usepackage{transparent}

\definecolor{njuPurple}{RGB}{220,205,230}     % 南大紫（深）
\definecolor{njuPurpleLight}{RGB}{250,245,252}   % 极浅的紫色背景（接近白）

\newtcolorbox{abstractbox}{
    colback=njuPurpleLight,   % 浅紫色背景
    colframe=njuPurple,       % 深紫色边框
    boxrule=1pt,              % 边框粗细
    arc=4mm,                  % 圆角
    left=8pt,                 % 左边距
    right=8pt,                % 右边距
    top=8pt,                  % 上边距
    bottom=8pt,               % 下边距
    opacityback=0.95
}

\title{CLI-Universe: Towards Verifiable Task Synthesis Engine for Terminal Agents
}

\author{
\textbf{Nanjing University} \quad \textbf{StepFun}  \quad
\textbf{ZODA}  \quad \textbf{Shanghai AI Lab} \\
\textbf{Huazhong University of Science and Technology}  \\[6pt]
\normalsize See \hyperref[sec:contributions]{Contributions section} for a full author list.
}

\begin{document}

\maketitle
\let\oldthefootnote\thefootnote

\let\thefootnote\relax\footnotetext{*~Equal Contribution. ~~$^\dagger$~Corresponding Author.}
\let\thefootnote\oldthefootnote

\begin{abstractbox}
\begin{center}
\textbf{\Large Abstract}
\end{center}
\noindent

While recent LLM-based terminal agents have demonstrated promising capabilities, the scarcity of high-quality, executable training data remains a critical bottleneck. Existing synthesis pipelines typically scale by retrofitting surface-level artifacts into tasks, frequently yielding ambiguous instructions, shallow execution paths, and brittle tests that provide weak learning signals. To overcome this, we introduce \textbf{CLI-Universe}, a principled synthesis engine that constructs terminal-agent tasks. CLI-Universe generates candidate tasks by sampling combinations across a multi-dimensional capability taxonomy (domain, skill type, capability, and engineering pillar), then grounds each candidate through evidence-guided deep research over real-world technical materials. To ensure rigorous supervision, validated blueprints are instantiated into Dockerized environments and subjected to a multi-stage executable verification pipeline—featuring rubric-gated test construction, hint-conditional filtering, and strict fail-to-pass checking. Across the full pipeline, from candidate generation to verification, approximately two-thirds of candidates are discarded, retaining only those that are genuine, verifiable, and non-trivially challenging. To validate our framework, we instantiate a highly distilled dataset of 6,000 trajectories called \textbf{CLI-Universe-6K}. Remarkably, fine-tuning Qwen3-32B on CLI-Universe-6K achieves 33.4\% on Terminal-Bench 2.0. This sets a new state-of-the-art for models trained on open-source data at or below 32B parameters, and outperforms several models an order of magnitude larger, demonstrating the profound data efficiency of structured, high-fidelity synthesis.

% Recent terminal agents have demonstrated strong performance on complex CLI tasks, but high-quality training data for building capable models remains scarce. Existing synthesis pipelines scale data by converting surface materials into tasks, but direct conversion often produces ambiguous queries, shallow execution, or brittle tests that provide weak learning signal. We introduce \textbf{CLI-Universe}, a pipeline that constructs terminal-agent tasks from structured capability specifications combined with evidence-guided deep research. Each task candidate is defined through a structured taxonomy combining domain, skill type, capability, and engineering pillar, then refined via deep research over real technical materials to ground it in realistic constraints and failure modes. Validated blueprints are instantiated into Dockerized environments, and a multi-stage executable verification pipeline applies rubric-gated test construction, hint-conditional filtering, and fail-to-pass checking to reject approximately two-thirds of candidates, retaining only tasks that are valid, verifiable, and non-trivially challenging. After fine-tuning Qwen3-32B on only 6k CLI-Universe trajectories, our model achieves \textbf{33.4\%} on Terminal-Bench~2, surpassing all models trained with open-source data at 32B or below and outperforming several models an order of magnitude larger on the official leaderboard.

\end{abstractbox}

\section{Introduction}

Recent terminal agents~\citep{claudecode, codexcli, geminicli, opencode2025, minisweagent2025, wang2025openhandsopenplatformai, merrill2026terminalbenchbenchmarkingagentshard} have demonstrated increasingly strong performance on complex CLI tasks, from software debugging and system administration to security analysis and data engineering. Yet high-quality training data for building capable terminal-agent models remains scarce. Effective training requires tasks that are genuinely difficult, demanding multi-step interaction with realistic environments, and validated by rigorous executable checks that provide unambiguous supervision signal rather than merely confirming that a script runs. Existing synthesis pipelines~\citep{wu2026largescaleterminalagentictrajectory, pi2026dataengineeringscalingllm, fan2026scalableterminaltasksynthesis, gandhi2026endlessterminalsscalingrl, lin2026cligymscalableclitask, zhu2026termigenhighfidelityenvironmentrobust} attempt to address this gap by converting readily available surface materials into tasks, for example by mining repositories and documentation, transforming execution traces and templates, or repurposing existing benchmarks. However, these materials were not originally designed as training tasks, and direct conversion often produces ambiguous queries, shallow execution paths, or brittle tests that provide weak learning signal even when they run successfully. In short, existing pipelines scale task sources rather than task quality.

To address this, we propose \textbf{CLI-Universe}, a pipeline that constructs terminal-agent tasks from structured capability specifications combined with evidence-guided deep research. Rather than retrofitting tasks from existing artifacts, CLI-Universe takes an inside-out approach: each task candidate is first defined through a structured taxonomy that combines a target domain with specific skill types, capabilities, and engineering pillars, ensuring broad and controlled coverage of the capability space. Each candidate is then refined through deep research over real technical materials: we search repositories, documentation, issue discussions, and usage examples to ground the task in realistic tools, constraints, and failure modes. This evidence-guided process transforms abstract capability specifications into technically grounded blueprints that exercise meaningful engineering workflows.

Validated blueprints are instantiated into Dockerized environments with materialized assets and runtime state. To ensure reliable supervision, CLI-Universe applies a multi-stage executable verification pipeline. First, role-separated agents independently construct rubric-gated tests and reference solutions, where the test agent and solution agent operate without seeing each other's output. Second, hint-conditional filtering retains only tasks where the internal hint is genuinely necessary for success, removing trivially solvable instances that would provide little training value. Finally, a fail-to-pass check provides executable proof that each retained task realizes a meaningful state transition from an unsolved state to a verified solution. Together, these stages reject approximately two-thirds of all candidates, concentrating supervision on high-signal examples.

After fine-tuning Qwen3-32B~\citep{yang2025qwen3technicalreport} on only 6k CLI-Universe trajectories, our model achieves \textbf{33.4\%} on Terminal-Bench~2~\citep{merrill2026terminalbenchbenchmarkingagentshard}, surpassing all models trained with open-source data at $\leq$32B scale and outperforming several open-weight models an order of magnitude larger on the official leaderboard. Performance scales monotonically with model size and generalizes to out-of-domain benchmarks, suggesting that structured specification, evidence-guided deep research, and multi-stage verification can together provide strong supervision even at limited data scale.

Our main contributions are:
\begin{itemize}
\setlength{\itemsep}{2pt}
\item \textbf{A data synthesis pipeline for terminal-agent training.} CLI-Universe defines each task through a structured capability taxonomy and refines it via evidence-guided deep research over real technical materials, producing Dockerized environments grounded in realistic constraints and failure modes. Unlike pipelines that retrofit tasks from surface artifacts, this inside-out design ensures that tasks exercise meaningful capability patterns by construction.

\item \textbf{Multi-stage verification that concentrates training signal.} CLI-Universe applies rubric-gated test construction, hint-conditional filtering, and fail-to-pass checking to reject approximately two-thirds of candidates, retaining only tasks with high-fidelity supervision. Ablation shows that removing any single component costs 3--6 points on Terminal-Bench~2, confirming that each verification stage drives downstream performance.

\item \textbf{A high-fidelity 6k-trajectory training set (CLI-Universe-6K) and empirical validation.} We construct CLI-Universe-6K, a 6{,}000-trajectory training set distilled from the pipeline. Fine-tuning Qwen3-32B on CLI-Universe-6K achieves 33.4\% on Terminal-Bench~2, surpassing all models trained with open-source data at $\leq$32B scale and outperforming several models an order of magnitude larger on the official leaderboard, demonstrating the data efficiency of our pipeline. Performance scales monotonically with model size and generalizes to out-of-domain benchmarks including BFCL~v4 and VitaBench~\citep{pmlr-v267-patil25a,he2025vitabench}.
\end{itemize}

\section{Related Work}

\paragraph{Terminal Agents.}
LLM-based agents have seen rapid progress in the code domain, with capabilities expanding from repository-level issue resolution~\citep{yang2024sweagentagentcomputerinterfacesenable, jimenez2024swebenchlanguagemodelsresolve} to interactive terminal environments where agents execute multi-step workflows through command-line interfaces. A growing collection of agent scaffolds~\citep{claudecode, codexcli, geminicli, wang2025openhandsopenplatformai, opencode2025} supports this paradigm by enhancing LLMs' planning, execution, and tool-calling capabilities. To evaluate terminal-agent performance, Terminal-Bench~\citep{merrill2026terminalbenchbenchmarkingagentshard} provides hand-crafted tasks spanning diverse domains within containerized Docker environments. However, open-source models remain substantially behind proprietary systems on these tasks, and performance improvements are increasingly bottlenecked by the scarcity of high-quality terminal-agent training data.

\paragraph{Synthetic Data for Terminal Agents.}
Existing approaches to scaling terminal-agent training data follow two main strategies. The first generates tasks from skill or domain taxonomies produced by LLMs~\citep{gandhi2026endlessterminalsscalingrl, zhu2026termigenhighfidelityenvironmentrobust, pi2026dataengineeringscalingllm, fan2026scalableterminaltasksynthesis}, achieving broad topical coverage by systematically enumerating capability dimensions. The second extracts tasks from existing infrastructure, e.g.\ by harvesting Docker environments from open-source repositories~\citep{wu2026largescaleterminalagentictrajectory} or transforming working configurations into faulty states to create debugging scenarios~\citep{lin2026cligymscalableclitask}. Both strategies effectively increase task volume, but offer limited guarantees on per-task verification quality and training signal density. \ours{} takes a complementary approach: it combines structured capability specification with evidence-guided deep research and multi-stage executable verification, so that each retained trajectory is grounded in realistic technical constraints and validated by rigorous executable checks.

\section{Method}
\label{sec:method}

\subsection{Overview}
\label{sec:method_overview}

\begin{figure*}[t]
    \centering
    \includegraphics[width=\textwidth,height=0.38\textheight,keepaspectratio]{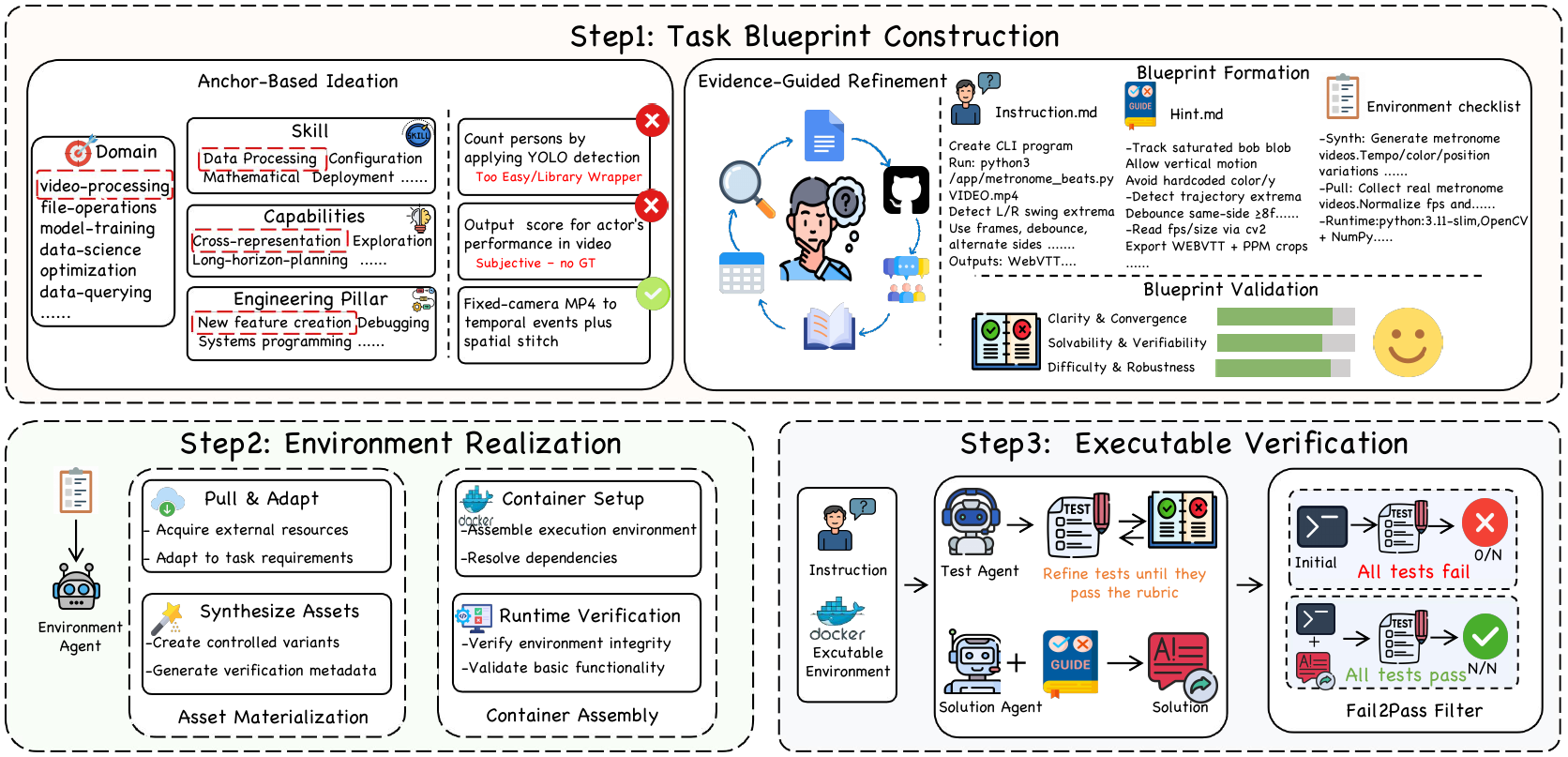}
    \caption{\textbf{Overview of CLI-Universe.}
    \emph{Step~1 (Query Construction):} A structured taxonomy seeds diverse task candidates; each is grounded through iterative deep research and compiled into a blueprint.
    \emph{Step~2 (Environment Synthesis):} Blueprints are realized into Docker containers with materialized assets and verified runtime.
    \emph{Step~3 (Validation \& Filtering):} A test agent and a solution agent independently verify the task; only instances passing rubric-gated tests and fail-to-pass checks are retained.}
    \label{fig:pipeline}
\end{figure*}

CLI-Universe constructs terminal-agent tasks through a structured synthesis pipeline (Figure~\ref{fig:pipeline}) with three stages. First, task blueprint construction (Section~\ref{sec:blueprint_construction}) specifies candidates along domain, skill type, capability, and engineering pillar dimensions, then iteratively refines each candidate through a research agent that searches real technical materials and incorporates evidence into the task specification, producing validated blueprints. Second, environment realization (Section~\ref{sec:environment_synthesis}) turns each blueprint into a fully executable Dockerized environment. Third, validation with executable filtering (Section~\ref{sec:test_construction_filtering}) constructs tests and reference solutions, retaining only candidates that pass fail-to-pass checking. Candidates are progressively filtered at each stage, and only 33.6\% survive end-to-end (Figure~\ref{fig:task_production_funnel}).

\subsection{Task Blueprint Construction}
\label{sec:blueprint_construction}

\paragraph{Task Candidate Specification.}
We anchor idea generation on three orthogonal dimensions beyond domain: (1)~\textbf{skill type}, the specialized technical knowledge required (e.g., algorithmic, systems, configuration, cryptography); (2)~\textbf{capability}, the reasoning behavior the task should elicit (e.g., exploration, error recovery, constraint satisfaction, long-horizon planning); and (3)~\textbf{engineering pillar}, the form of work the agent performs (e.g., new feature creation, debugging, DevOps, refactoring). For each domain, we predefine a pool of values per dimension and sample combinations as anchor points. Given an anchor, we brainstorm task ideas constrained to exercise the specified capability pattern, ensuring diversity at the skill level rather than only in surface topic. Generated ideas are scored for creativity, technical grounding, and feasibility, and only top-scoring candidates proceed to evidence-guided refinement. The full taxonomy is given in Appendix~\ref{sec:task-taxonomy}.

\begin{figure*}[t]
    \centering
    \begin{subfigure}[t]{0.37\textwidth}
        \centering
        \includegraphics[width=\linewidth]{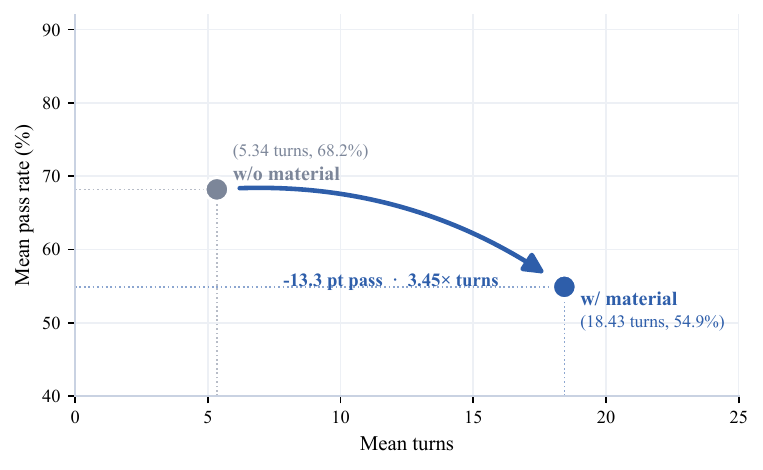}
        \caption{Solver turns and pass rate, before vs.\ after refinement.}
        \label{fig:material_effect}
    \end{subfigure}
    \hfill
    \begin{subfigure}[t]{0.37\textwidth}
        \centering
        \includegraphics[width=\linewidth]{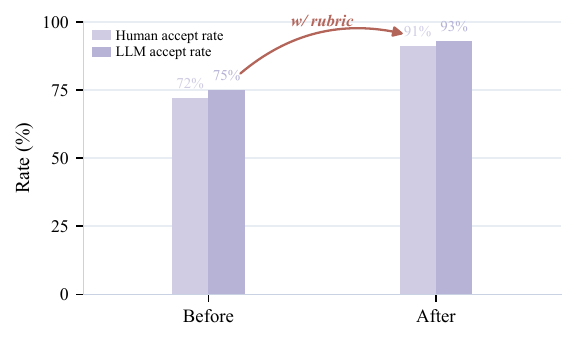}
        \caption{Blueprint accept rate, before vs.\ after rubric review.}
        \label{fig:blueprint_refinement}
    \end{subfigure}
    \hfill
    \begin{subfigure}[t]{0.24\textwidth}
        \centering
        \includegraphics[width=\linewidth]{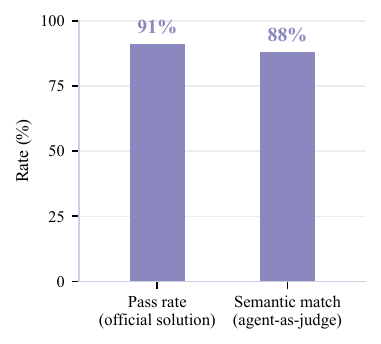}
        \caption{Agreement between our tests and TB-2 ground truth.}
        \label{fig:test_consistency}
    \end{subfigure}\\[6pt]
    \begin{subfigure}[t]{0.55\textwidth}
        \centering
        \includegraphics[width=\linewidth]{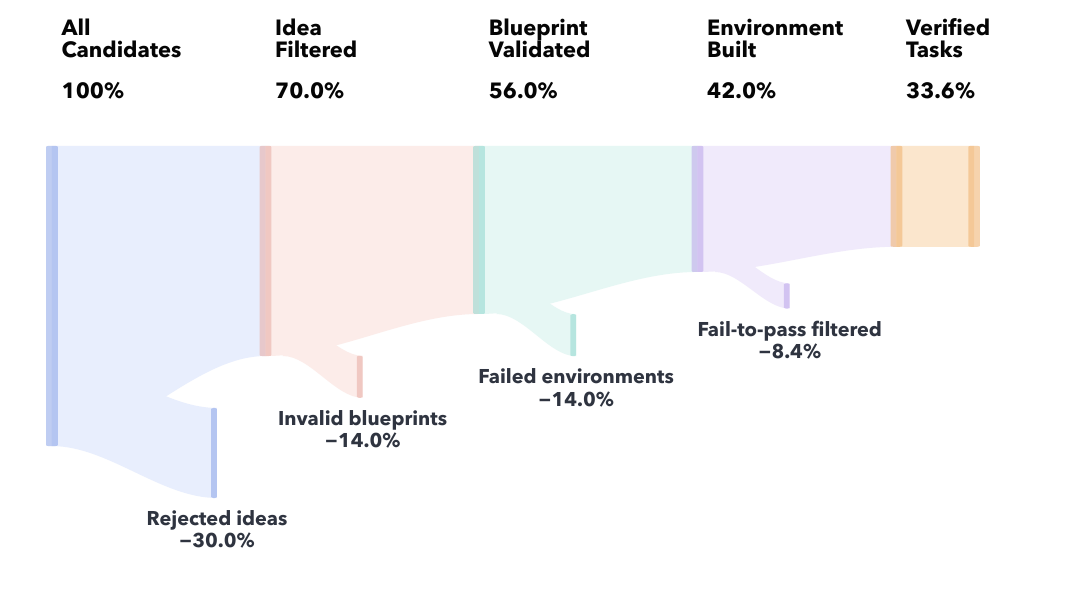}
        \caption{Per-stage retention across the five filters.}
        \label{fig:task_production_funnel}
    \end{subfigure}
    \hspace{6pt}
    \begin{subfigure}[t]{0.42\textwidth}
        \centering
        \includegraphics[width=\linewidth]{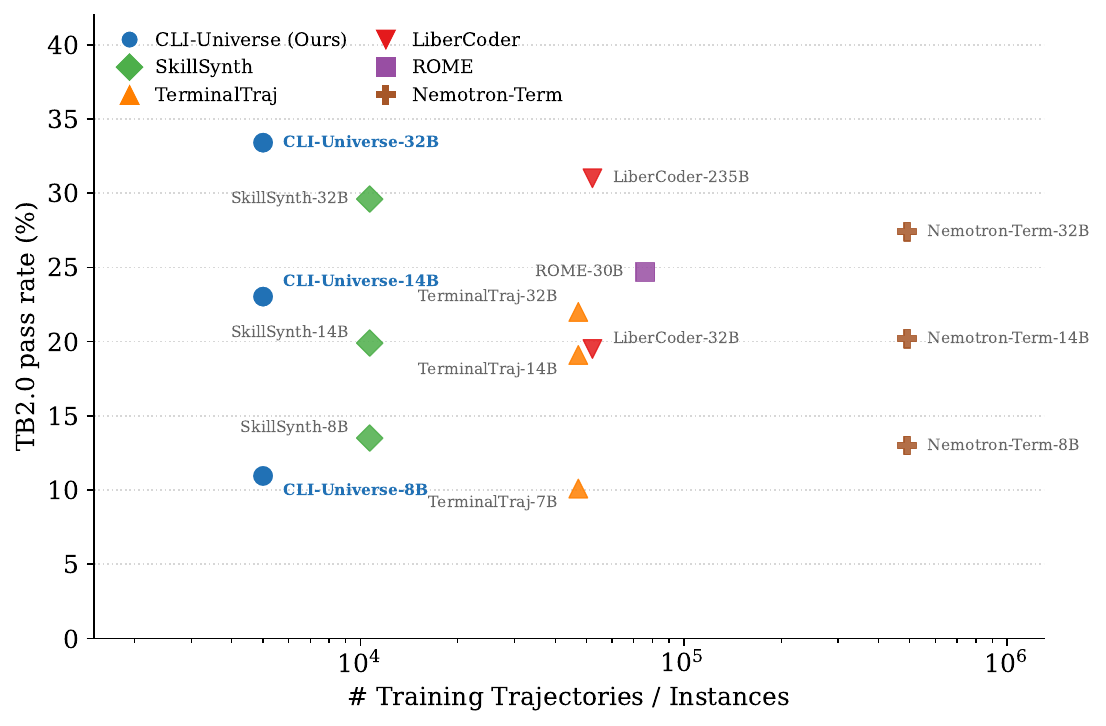}
        \caption{TB-2.0 score vs.\ number of training trajectories.}
        \label{fig:data_scale_comparison}
    \end{subfigure}
    \caption{\textbf{Each synthesis stage adds a measurable gain, and the resulting dataset trains agents with $10$--$100\times$ fewer trajectories.} (a)--(c) validate the three pipeline stages: evidence refinement raises difficulty ($3.45\times$ solver turns, $-13.3$ pt pass rate); rubric-based blueprint review lifts accept rate to $91\%$/$93\%$ (human/LLM); synthesized tests agree with Terminal-Bench~2 ground truth on $91\%$ of sampled tasks ($88\%$ semantic match, Codex / GPT-5.4 judge). (d) The five-stage filter retains $33.6\%$ of candidates end-to-end. (e) Models trained on CLI-Universe data match or exceed TB~2.0 baselines using $1$--$2$ orders of magnitude fewer training trajectories.}
    \label{fig:method_validation}
\end{figure*}

\paragraph{Evidence-Guided Refinement.}
Each task candidate undergoes iterative refinement driven by a dedicated research agent. Starting from the abstract idea, the agent searches real technical materials (repositories, documentation, issue discussions, tutorials, and usage examples) and progressively incorporates the evidence into the task specification, grounding it in specific tools, realistic constraints, known failure modes, and concrete input/output contracts. The agent continues refining until the specification is sufficiently complete to support query formulation, environment construction, and test generation. Candidates that cannot be adequately grounded (e.g., those requiring unavailable tools or conflicting constraints) are discarded.
As shown in Figure~\ref{fig:material_effect}, tasks with evidence-guided refinement require $3.45\times$ more solver turns and reduce pass rate by $13.3$ points compared to unrefined tasks, confirming that the refinement raises genuine difficulty rather than merely lengthening trajectories.

\paragraph{Blueprint Formation and Validation.}
Each refined candidate is compiled into a blueprint that records the user-facing instruction, an internal hint for reference-solution construction, and an environment checklist. CLI-Universe validates each blueprint before proceeding, retaining only those whose specification is sufficiently clear and whose setup admits reliable downstream verification. As shown in Figure~\ref{fig:blueprint_refinement}, human and LLM judges show high agreement, and rubric-based validation substantially increases acceptance rates (human: 72\% $\to$ 91\%; LLM: 75\% $\to$ 93\%).

\subsection{Environment Realization}
\label{sec:environment_synthesis}

Only validated blueprints are carried forward to environment realization, where CLI-Universe turns each blueprint into a fully executable Dockerized environment.

\paragraph{Asset Materialization.}
Guided by the environment checklist in the blueprint, CLI-Universe acquires the required assets (source repositories, documentation, datasets, configuration files, service logs, etc.) from publicly available resources. Retrieved assets rarely match the task specification exactly, so the system adapts them to fit: normalizing data formats, injecting controlled faults, adjusting configuration parameters, or scoping content to the intended workflow. When suitable external assets are unavailable, the system synthesizes them from scratch, creating controlled variants with known ground-truth properties and generating the verification metadata needed for downstream testing.

\paragraph{Environment Assembly.}
The materialized assets are then packaged into a Docker image together with the required dependencies, runtime configuration (environment variables, services, permissions), and pinned package versions to ensure reproducibility. The system places each asset at the filesystem location specified by the blueprint and wires up any inter-component references (e.g., paths in configuration files, database connection strings) so that the environment is self-contained. Each assembled environment undergoes a suite of smoke tests that verify successful dependency installation, correct service startup, expected filesystem layout, and basic end-to-end reachability. Environments that fail these checks are discarded.

\subsection{Test Construction and Executable Filtering}
\label{sec:test_construction_filtering}

CLI-Universe constructs both executable tests and a reference solution trajectory for each task. The tests operationalize task completion with respect to the user-facing query, while the solution trajectory provides a successful demonstration under internal guidance and is retained as training data for the final dataset.

\paragraph{Test Construction.}
CLI-Universe first constructs task-specific tests from the realized environment and the validation target specified in the blueprint. The test agent generates a candidate test suite, then iteratively checks each test against a set of test-case rubrics covering correctness, determinism, and edge-case coverage, refining or replacing tests until they provide a stable executable signal for the task.

To assess the validity of the constructed tests, we apply the same test-construction pipeline to 89 Terminal-Bench2 tasks and compare the resulting test suites with the official ones. As shown in Figure~\ref{fig:test_consistency}, the official solutions pass our synthesized tests on $91\%$ of sampled tasks, and agent-as-judge semantic match (using Codex with GPT-5.4) between our tests and the official ones reaches $88\%$. These results indicate that our pipeline produces tests that are consistent with the official ground truth.

\paragraph{Solution Construction.}
A solution agent is prompted with the realized environment and an internal hint from the blueprint to produce a successful solution trajectory. The hint provides key resolution steps and expected intermediate states without appearing in the actual task query, anchoring an intended resolution path while preserving the external difficulty of the task. The resulting trajectory is retained as training data only when it resolves the task under the realized environment and remains consistent with the executable completion signal established by the test suite.

\paragraph{Hint-Conditional Filtering.}
We further filter tasks and solutions by comparing solution attempts with and without the internal hint. A separate agent attempts each task without access to the hint; a candidate is retained only when the hint-free attempt fails while the hint-guided attempt succeeds. This removes trivially solvable instances so that the final dataset concentrates on task--trajectory pairs where the supervision provides meaningful training value.

\paragraph{Fail-to-Pass Filtering.}
A final fail-to-pass condition is imposed on each retained solution. The generated tests must fail on the initial environment and pass after execution of the hinted solution trajectory. This bidirectional check removes both vacuous tasks whose tests pass trivially and unsupported trajectories that do not actually reach the goal state, ensuring that every retained example realizes an executable transition from an unsolved state to a verified solution.

\section{Experiments}
\label{sec:experiments}

We use the Qwen3~\citep{yang2025qwen3technicalreport} dense series (8B, 14B, and 32B) as base models and fine-tune each on 6k CLI-Universe trajectories collected from Kimi-K2.6~\citep{kimiteam2026kimik2openagentic}. Full training details are in Appendix~\ref{sec:training_details}. We evaluate on Terminal-Bench~1.0 and 2.0~\citep{merrill2026terminalbenchbenchmarkingagentshard} (TB~1.0, TB~2.0) mainly using the Terminus 2 scaffold with 200 turns per task and report avg@4. Generalization is measured on BFCL~v4~\citep{pmlr-v267-patil25a} and VitaBench~\citep{he2025vitabench}.
% We use the Qwen3~\citep{yang2025qwen3technicalreport} dense series (8B, 14B, and 32B) as base models and fine-tune each on 6k CLI-Universe trajectories collected from Kimi K2.6~\citep{kimiteam2026kimik2openagentic}. Full training details are in Appendix~\ref{sec:training_details}. We evaluate on Terminal-Bench~1.0 and 2.0~\citep{merrill2026terminalbenchbenchmarkingagentshard} (TB~1.0, TB~2.0) mainly using the Terminus2 scaffold with 200 turns per task and report avg@4. Generalization is measured on BFCL~v4~\citep{pmlr-v267-patil25a} and VitaBench~\citep{he2025vitabench}.

\FloatBarrier
\subsection{Main Results}
\label{sec:main_results}

\paragraph{CLI-Universe surpasses all models trained with open-source data at $\leq$32B.} Table~\ref{tab:main_results} reports performance on Terminal-Bench~1.0 and 2.0 under the Terminus~2 scaffold. CLI-Universe-32B reaches 33.4 on TB~2.0, outperforming SkillSynth-32B~\citep{fan2026scalableterminaltasksynthesis} (29.6), Nemotron-Terminal-32B~\citep{pi2026dataengineeringscalingllm} (27.4), and TerminalTraj-32B~\citep{wu2026largescaleterminalagentictrajectory} (22.0). The advantage holds at 14B and 8B. Our 32B model also surpasses several open-weight systems an order of magnitude larger (e.g., 480B Qwen3-Coder at 23.9, 1T Kimi-K2-Instruct at 27.8), though a gap to the strongest proprietary systems (e.g., Claude-Opus-4.5 at 57.8) remains.

\paragraph{Gains scale with model size.} Figure~\ref{fig:model_scaling} reports TB~2.0 scores of Qwen3 baselines and our CLI-Universe models at three sizes. We observe two trends. First, the Qwen3 baselines stay essentially flat across scales (2.5 / 4.0 / 3.4 at 8B / 14B / 32B); without targeted agentic data, simply enlarging the base model does \emph{not} unlock terminal-agent capability. Second, after training on CLI-Universe trajectories the same checkpoints jump to 10.9 / 23.0 / 33.4, and the gain over baseline grows monotonically with model size ($+8.4 \to +19.0 \to +30.0$). This suggests that CLI-Universe data scales with model capacity rather than saturating, and that larger students extract more value from the same teacher trajectories, a regime where bigger models are not yet bottlenecked by data difficulty.

\begin{table*}[t]
\centering
\caption{\textbf{Main results on Terminal-Bench 1.0 and 2.0.} Models are sorted by TB~2.0 within each group. CLI-Universe models surpass all models trained with open-source trajectory data at comparable size ($\leq$32B). All baselines use the Terminus~2 scaffold except LiberCoder, which uses OpenHands.}
\label{tab:main_results}
\footnotesize
\setlength{\tabcolsep}{6pt}
\renewcommand{\arraystretch}{1.0}
\begin{tabular}{lccc}
\toprule
\textbf{Model} & \textbf{Size} & \textbf{Terminal-Bench 1.0} & \textbf{Terminal-Bench 2.0} \\
\midrule
\multicolumn{4}{c}{\textbf{Closed-Source Models}} \\
\midrule
Claude-Opus-4.5~\citep{claude4modelcard}        & --   & 47.5 & 57.8 \\
Gemini 3 Pro Preview~\citep{gemini25}           & --   & 46.4 & 56.9 \\
GPT-5.2~\citep{gpt5}                            & --   & 54.4 & 54.0 \\
\midrule
\multicolumn{4}{c}{\textbf{Open-Source Models}} \\
\midrule
GLM-4.7~\citep{glm5team2026glm5vibecodingagentic}                   & 358B & 48.8 & 41.0 \\
LiberCoder-235B~\citep{lin2026cligymscalableclitask}                & 235B & 46.1 & 31.0 \\
SkillSynth-32B~\citep{fan2026scalableterminaltasksynthesis}         & 32B  & 33.8 & 29.6 \\
Minimax-M2.1~\citep{minimax2026minimaxm2seriesminiactivations}      & 229B & 42.5 & 29.2 \\
Kimi-K2-Instruct~\citep{kimiteam2026kimik2openagentic}              & 1T   & 44.6 & 27.8 \\
Nemotron-Terminal-32B~\citep{pi2026dataengineeringscalingllm}       & 32B  & --   & 27.4 \\
Qwen3-Coder-480B~\citep{Qwen3-Coder-Next}                           & 480B & 37.5 & 23.9 \\
TerminalTraj-32B~\citep{wu2026largescaleterminalagentictrajectory}  & 32B  & 35.3 & 22.0 \\
Nemotron-Terminal-14B~\citep{pi2026dataengineeringscalingllm}       & 14B  & --   & 20.2 \\
SkillSynth-14B~\citep{fan2026scalableterminaltasksynthesis}         & 14B  & 22.9 & 19.9 \\
LiberCoder-32B~\citep{lin2026cligymscalableclitask}                 & 32B  & 38.9 & 19.5 \\
TerminalTraj-14B~\citep{wu2026largescaleterminalagentictrajectory}  & 14B  & 28.9 & 19.1 \\
Qwen3-Coder-30B-A3B~\citep{Qwen3-Coder-Next}                        & 30B  & 23.8 & 14.6 \\
SkillSynth-8B~\citep{fan2026scalableterminaltasksynthesis}          & 8B   & 17.1 & 13.5 \\
Nemotron-Terminal-8B~\citep{pi2026dataengineeringscalingllm}        & 8B   & --   & 13.0 \\
TerminalTraj-7B~\citep{wu2026largescaleterminalagentictrajectory}   & 7B   & 23.0 & 10.1 \\
Qwen3-14B~\citep{yang2025qwen3technicalreport}                      & 14B  & 11.6 &  4.0 \\
Qwen3-32B~\citep{yang2025qwen3technicalreport}                      & 32B  & 11.2 &  3.4 \\
Qwen3-8B~\citep{yang2025qwen3technicalreport}                       & 8B   &  8.1 &  2.5 \\
\midrule
\multicolumn{4}{c}{\textbf{Ours}} \\
\midrule
\textbf{CLI-Universe-32B} & 32B & \textbf{38.8} & \textbf{33.4} \\
\textbf{CLI-Universe-14B} & 14B & \textbf{27.5} & \textbf{23.0} \\
\textbf{CLI-Universe-8B}  & 8B  & \textbf{19.1} & \textbf{10.9} \\
\bottomrule
\end{tabular}
\end{table*}

\FloatBarrier
\subsection{Ablation Studies}
\label{sec:ablations}

We isolate the contribution of three pipeline components and two data-side choices (trajectory selection and teacher model). Unless noted, the student model is Qwen3-32B, and we report TB~2.0 scores.

\begin{table}[!ht]
\centering
\caption{\textbf{Data-side ablations on Qwen3-32B.} (a) trajectory selection strategy; (b) teacher model choice. TB~2.0 scores reported as avg@4.}
\label{tab:ablation_data}
\small
\begin{subtable}[t]{0.48\linewidth}
\centering
\setlength{\tabcolsep}{8pt}
\caption{Trajectory selection.}
\label{tab:ablation_selection}
\begin{tabular}{l c c}
\toprule
\textbf{Strategy} & \textbf{\#Traj.} & \textbf{TB2.0} \\
\midrule
Complete (all kept) & 10k & 28.2 \\
Success-only        & 6k  & \textbf{33.4} \\
\bottomrule
\end{tabular}
\end{subtable}
\hfill
\begin{subtable}[t]{0.48\linewidth}
\centering
\setlength{\tabcolsep}{8pt}
\caption{Teacher model.}
\label{tab:ablation_teachers}
\begin{tabular}{l c c}
\toprule
\textbf{Teacher} & \textbf{\#Traj.} & \textbf{TB2.0} \\
\midrule
DeepSeek-V4-Pro & 6k & 31.2 \\
Kimi-K2.6       & 6k & \textbf{33.4} \\
\bottomrule
\end{tabular}
\end{subtable}
\end{table}

\subsubsection{Effect of different components}
\label{sec:ablation_components_subsec}

Figure~\ref{fig:analysis_panels}(a) isolates the contribution of each pipeline component by removing it in turn while keeping the rest of the pipeline and training recipe fixed. To make the ablation study tractable, we conduct it on a 1k-task subset using the Qwen3-32B student. All three ablations cause a substantial drop relative to the full pipeline (26.7): removing the asset strategy is the most damaging ($-6.2$, down to 20.5), indicating that the diversity of seeded environments is a major driver of task coverage. Removing the query rubrics costs 3.4 points (down to 23.3), showing that even when assets and tests are in place, query quality continues to bound what the student can learn. Removing the test-case rubrics costs 3.9 points (down to 22.8), confirming that high-fidelity verification contributes meaningfully to trainable trajectories. Together these results suggest that the three components are complementary rather than redundant, and that quality control at each stage contributes to downstream performance.

\subsubsection{Effect of different selection strategies}
\label{sec:ablation_selection_subsec}

Table~\ref{tab:ablation_selection} compares trajectory filtering strategies on Qwen3-32B. Keeping only the 6k successful trajectories (those that pass all test cases) yields the best downstream performance (33.4 on TB~2.0), outperforming the unfiltered complete set of 10k trajectories (28.2) by $+5.2$ points. This suggests that failed and incomplete interactions introduce noise that degrades training signal at this model scale, and that quality filtering on correctness is more important than raw trajectory volume.

\subsubsection{Effect of different teachers}

Table~\ref{tab:ablation_teachers} shows the impact of teacher model choice on student performance. We sample 6k trajectories from each teacher on the same CLI-Universe tasks and distill into Qwen3-32B. Kimi-K2.6~\citep{kimiteam2026kimik2openagentic} reaches 33.4 on TB~2.0 while DeepSeek-V4-Pro~\citep{deepseek2026v4} scores 31.2, suggesting that the pipeline is robust to teacher choice and does not require a specific frontier model to produce strong supervision.

\FloatBarrier
\subsection{Scaling Analysis}
\label{sec:scaling_analysis}

\begin{figure*}[!t]
    \centering
    \begin{subfigure}[t]{0.32\textwidth}
        \centering
        \includegraphics[width=\linewidth]{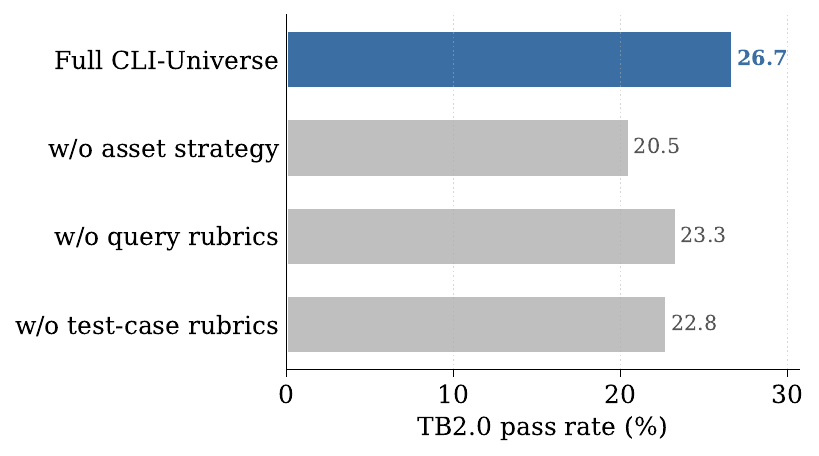}
        \caption{Component ablation.}
        \label{fig:ablation_components}
    \end{subfigure}
    \hfill
    \begin{subfigure}[t]{0.32\textwidth}
        \centering
        \includegraphics[width=\linewidth]{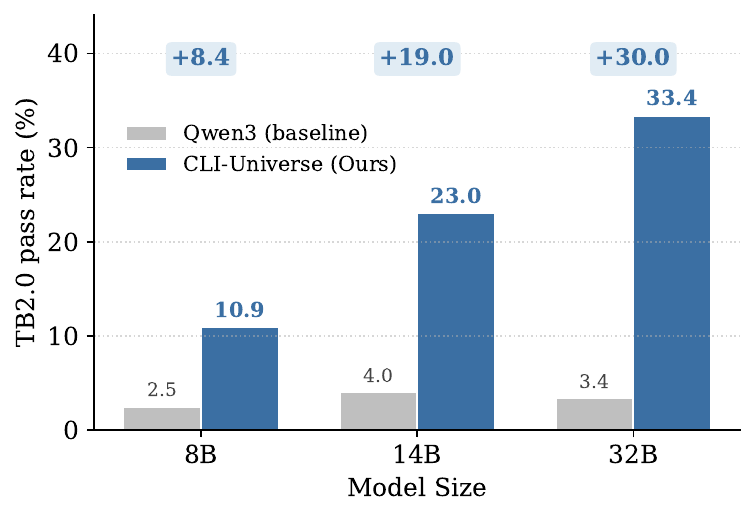}
        \caption{Model scaling.}
        \label{fig:model_scaling}
    \end{subfigure}
    \hfill
    \begin{subfigure}[t]{0.32\textwidth}
        \centering
        \includegraphics[width=\linewidth]{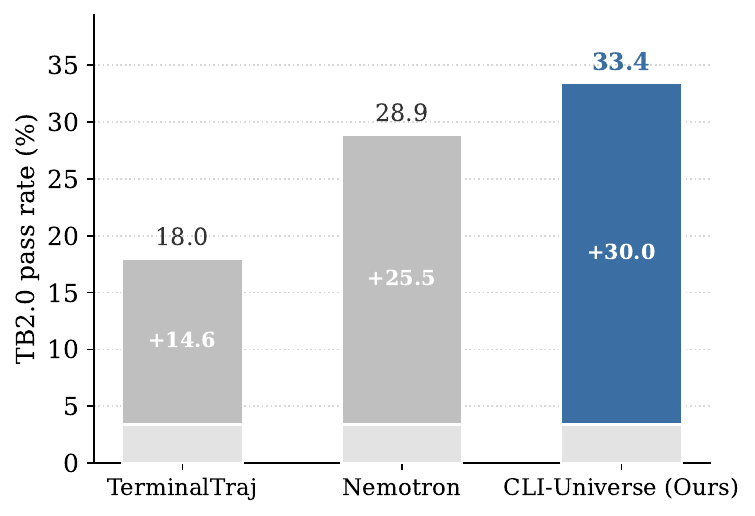}
        \caption{Data efficiency.}
        \label{fig:data_efficiency}
    \end{subfigure}
    \caption{\textbf{Ablation, scaling, and data efficiency of CLI-Universe.} (a) TB~2.0 score when each pipeline component is removed on a 1k-task subset using Qwen3-32B. (b) Qwen3 baselines vs.\ CLI-Universe at 8B / 14B / 32B on TB~2.0; blue badges show absolute gains. (c) TB~2.0 score of Qwen3-32B fine-tuned on 6k trajectories from each of TerminalTraj, Nemotron, and CLI-Universe (matched data volume); white labels show gain over the Qwen3-32B baseline.}
    \label{fig:analysis_panels}
\end{figure*}

\subsubsection{Data scaling}
\label{sec:analysis_data_scaling}

We investigate how the number of CLI-Universe training trajectories affects downstream performance. We train Qwen3-8B on subsets of increasing size and evaluate on TB~2.0. As shown in Figure~\ref{fig:data_scale_comparison}, performance improves steadily as data scales, with no sign of saturation at our largest budget. This suggests that the pipeline can continue to generate useful supervision and that the current data regime has not yet reached the point of diminishing returns.

\subsubsection{Data efficiency}
\label{sec:analysis_data_efficiency}

Figure~\ref{fig:analysis_panels}(c) compares the three data sources on Qwen3-32B. CLI-Universe achieves the highest TB~2.0 score (33.4), outperforming Nemotron~\citep{pi2026dataengineeringscalingllm} (28.9) and TerminalTraj~\citep{wu2026largescaleterminalagentictrajectory} (18.0). Measured as gain over the Qwen3-32B baseline (3.4), CLI-Universe yields $+30.0$ points, versus $+25.5$ for Nemotron and $+14.6$ for TerminalTraj. This suggests that the per-trajectory information density of CLI-Universe is markedly higher, and that benchmark gains here are driven by the quality of the underlying tasks and verification rather than trajectory count alone.

\FloatBarrier
\subsection{Generalization}
\label{sec:generalization}

\begin{figure*}[t]
    \centering
    \begin{subfigure}[t]{0.30\textwidth}
        \centering
        \includegraphics[width=\linewidth]{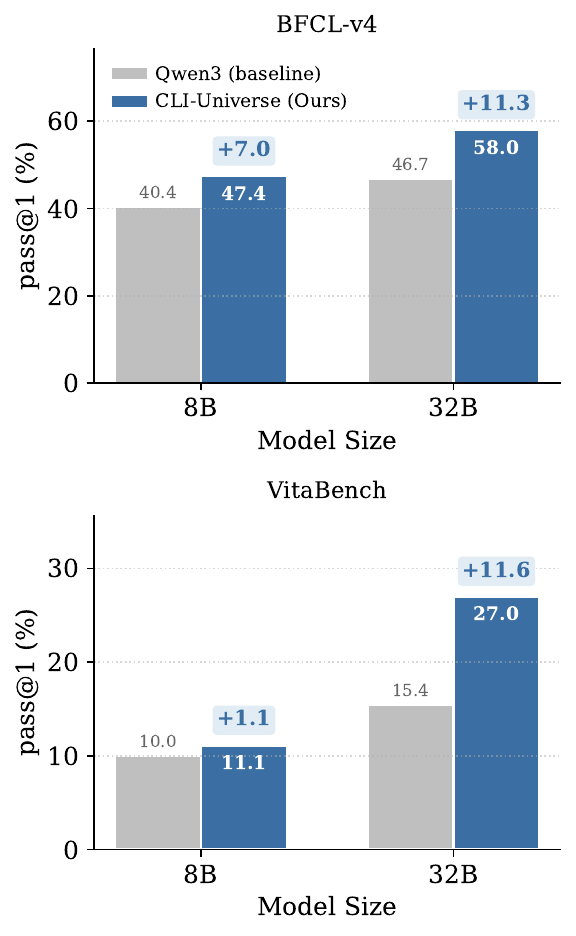}
        \caption{Generalization across agentic benchmarks (BFCL~v4, VitaBench).}
        \label{fig:generalization}
    \end{subfigure}
    \hfill
    \begin{subfigure}[t]{0.68\textwidth}
        \centering
        \includegraphics[width=\linewidth]{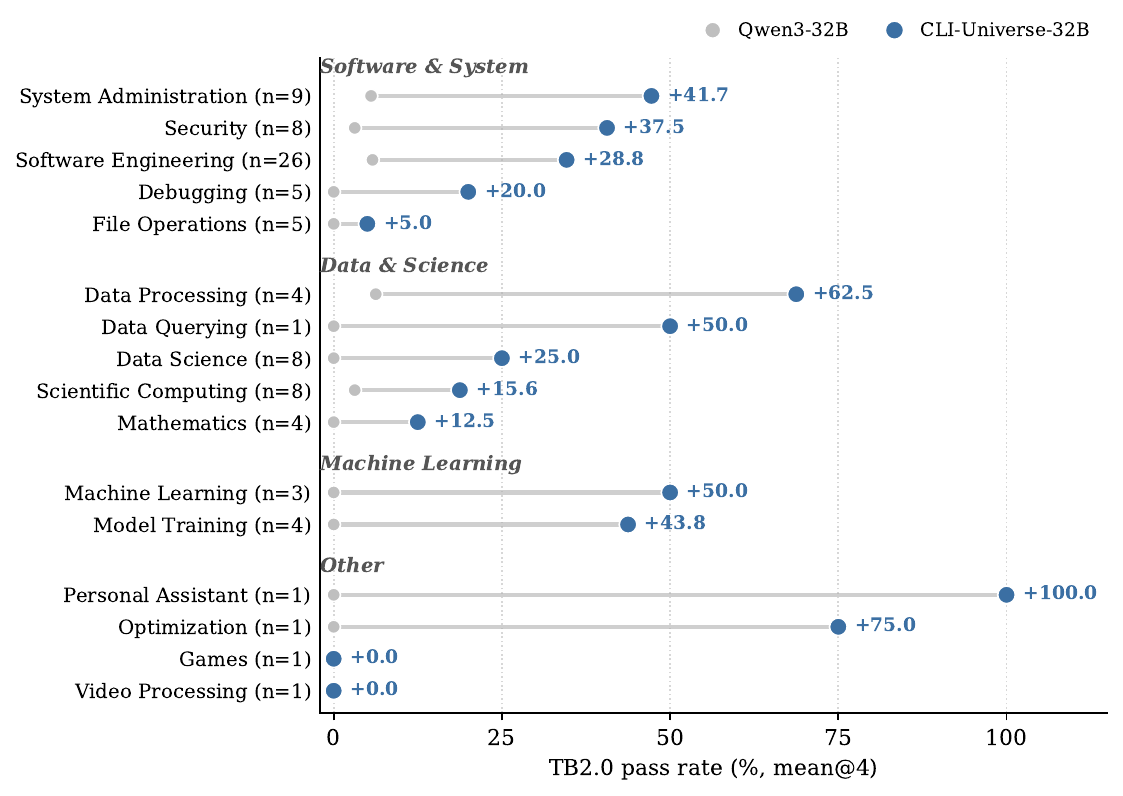}
        \caption{TB~2.0 pass rate by fine-grained category at 32B (avg@4).}
        \label{fig:category_breakdown}
    \end{subfigure}
    \caption{\textbf{Cross-benchmark and per-category performance at 32B.} (a) Qwen3-32B baselines vs.\ CLI-Universe-32B on two out-of-domain agentic benchmarks. (b) TB~2.0 avg@4 broken down by fine-grained categories grouped into four super-categories; gray dots show Qwen3-32B baseline.}
    \label{fig:gen_and_category}
\end{figure*}

\subsubsection{Cross-benchmark transfer}
\label{sec:analysis_generalization}

To verify that CLI-Universe training generalizes beyond Terminal-Bench, we evaluate on two additional agentic benchmarks: BFCL~v4~\citep{pmlr-v267-patil25a} (function calling) and VitaBench~\citep{he2025vitabench} (multi-turn tool-use). Figure~\ref{fig:generalization} shows consistent improvements: CLI-Universe-32B outperforms Qwen3-32B by $+11.3$ on BFCL~v4 (58.0 vs.\ 46.7) and $+11.6$ on VitaBench (27.0 vs.\ 15.4), with gains of $+7.0$ and $+1.1$ at 8B respectively. This cross-benchmark transfer suggests that the skills acquired from CLI-Universe (tool orchestration, environment state tracking, and multi-step planning) are broadly applicable rather than narrowly fitted to Terminal-Bench.

\subsubsection{Performance on different categories}
\label{sec:analysis_categories}

Figure~\ref{fig:category_breakdown} breaks down TB~2.0 avg@4 by fine-grained categories at 32B. The Qwen3-32B baseline scores near zero in most categories, whereas CLI-Universe training unlocks substantial capability across the board. The largest absolute gains appear in Data Processing ($+62.5$), Machine Learning ($+50.0$), Data Querying ($+50.0$), and Model Training ($+43.8$). In the Software \& System group, System Administration ($+41.7$), Security ($+37.5$), and Software Engineering ($+28.8$) all show strong improvements. A few categories remain challenging: Video Processing and Games show no gain at 32B, suggesting directions for future pipeline expansion.

\FloatBarrier
\subsection{Error Study}
\label{subsec:error-study}

\begin{figure}[!t]
\centering
\includegraphics[width=\linewidth]{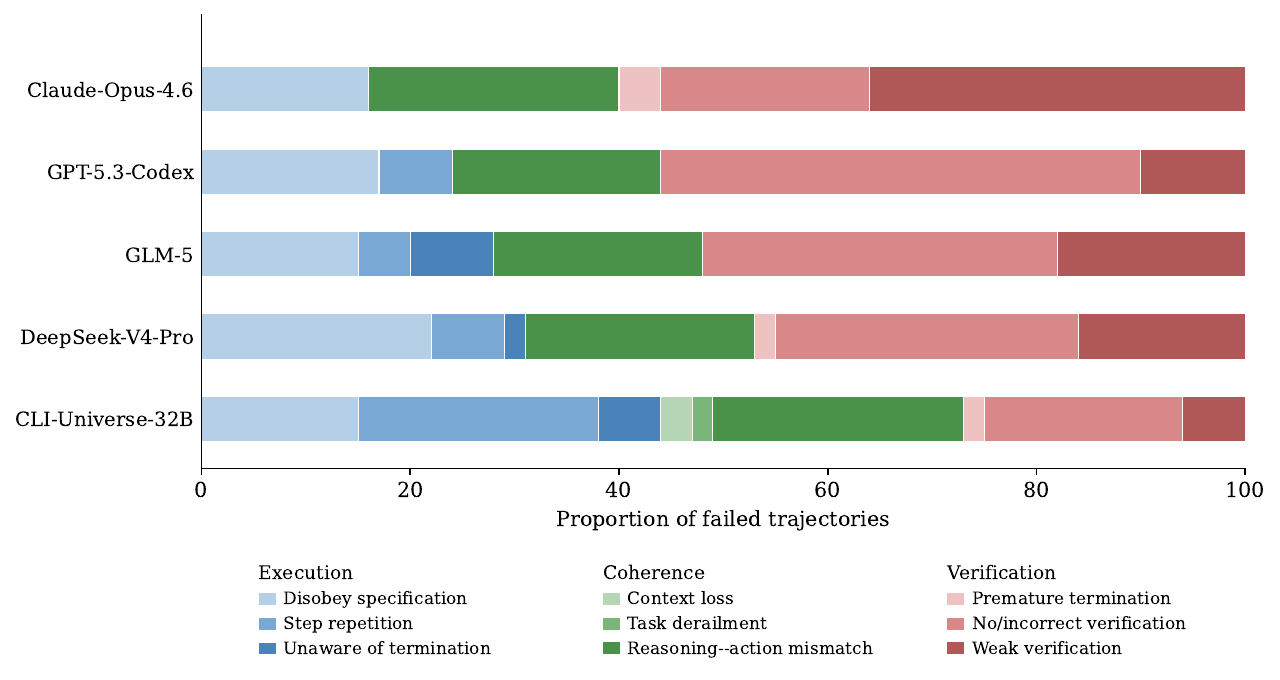}
\caption{\textbf{Primary failure attribution on failed TB~2.0 trajectories.} Each failed trajectory is assigned the single failure mode most causally responsible for the task failure (mutually exclusive, rows sum to 100\%). The 9 modes are organized into three classes: \emph{Execution} (disobey spec, step repeat, unaware of termination), \emph{Coherence} (context loss, derailment, reasoning--action mismatch), and \emph{Verification} (premature termination, no/incorrect verification, weak verification).}
\label{fig:error_study}
\end{figure}

We run two trajectory rollouts per task on Terminal-Bench~2 for each model, and use Codex with GPT-5.4 to classify each failed trajectory by the single failure mode most causally responsible for its failure, against a 9-mode taxonomy organized into three classes (Execution / Coherence / Verification, see Figure~\ref{fig:error_study}).

\paragraph{Frontier SOTA models fail mostly on the verification side.}
For all four proprietary baselines (Claude-Opus-4.6~\citep{claude4modelcard}, GPT-5.3-Codex~\citep{gpt5}, GLM-5~\citep{glm5team2026glm5vibecodingagentic}, and DeepSeek-V4-Pro~\citep{deepseek2026v4}), the \emph{Verification} class accounts for the largest share of failures, ranging from 47\% to 60\%. The pattern suggests these failures are not primarily an inability to execute or reason about the task: the agent typically makes a plausible attempt, but does not correctly check whether the goal state is actually satisfied before terminating.

\paragraph{Frontier SOTA split into two distinct verification styles.}
The Verification-side failures of SOTA fall into two opposite sub-modes. Opus is dominated by \emph{weak verification} (36\% vs.\ GPT's 10\%): it performs a check, but the check is too shallow to catch the failure. GPT is instead dominated by \emph{no/incorrect verification} (47\% vs.\ Opus's 20\%), as it tends to skip the check altogether. GLM-5 and DeepSeek-V4-Pro track the GPT pattern, with somewhat noisier execution (Execution share 28--31\%, against Opus's 16\% and GPT's 23\%).

\paragraph{CLI-Universe-32B's failure profile shifts to the execution side.}
For CLI-Universe-32B, the \emph{Verification} share drops to 27\% while \emph{Execution} becomes the largest class at 44\%. The most prominent shift is \emph{step repetition}, which rises from 0--7\% in the frontier baselines to 23\%. The failure mode is qualitatively different from the SOTA pattern: rather than skipping verification, the agent more often gets stuck during execution, looping on the same operation or failing to make stable progress on the task.

\section{Conclusion}
\label{sec:conclusion}

We presented CLI-Universe, a pipeline that constructs terminal-agent training
tasks through structured capability specification, evidence-guided deep
research, and multi-stage executable verification. Fine-tuning Qwen3 on the
resulting trajectories yields consistent gains across 8B, 14B, and 32B
students, and transfers to out-of-domain benchmarks including BFCL-v4 and
VitaBench. These results suggest that carefully constructed and independently
verified tasks can provide strong supervision for terminal agents even at
limited data scale.

\section{Limitations}
\label{sec:limitations}

Several aspects of this work deserve further scrutiny.
The pipeline relies on LLM-based agents for ideation, environment construction,
solution generation, and test construction; despite rubric gating and executable
verification, the quality ceiling of the synthesized data is ultimately bounded
by the capability of the underlying models.
Although CLI-Universe substantially narrows the gap to proprietary systems at
$\leq$32B scale, a clear performance gap to the strongest frontier models
remains, and we have not yet explored whether larger open-source base models or
reinforcement-learning-based training can further close this gap.
Finally, although CLI-Universe demonstrates strong data efficiency, the
current dataset contains only 6k trajectories; scaling the pipeline to
larger and more diverse task pools may unlock further gains, and we leave
this exploration to future work.

\phantomsection
\section*{Contributions}
\label{sec:contributions}

\noindent\textbf{Authors}~~Zhanbo Hua$^{3,*}$, Yifan Yao$^{1,*}$, Weihao Xie$^{4,*}$, Yongchi Zhao$^{2,*}$, Minghao Liu$^{3,*}$, Ruizhi Qiu$^{1}$, Zhewei Huang$^{2}$, Zun Wang$^{5}$, Yiyan Ji$^{1}$, Yunhai Ye$^{3}$, Letian Zhu$^{1}$, Xinping Lei$^{1}$, Han Li$^{1}$, Zhiyuan Ma$^{4}$, Zili Wang$^{2}$, Zhaoxiang Zhang$^{1}$, Jiaheng Liu$^{1,\dagger}$

\vspace{4pt}
\noindent\textbf{Affiliations}~~$^{1}$Nanjing University, $^{2}$StepFun, $^{3}$ZODA, $^{4}$Huazhong University of Science and Technology, $^{5}$Shanghai AI Lab
% , $^{6}$Beihang University

\vspace{4pt}
\noindent$^{*}$~Equal contribution.\\
$^{\dagger}$~Corresponding author.

\bibliographystyle{unsrtnat}
\bibliography{reference}

\appendix

\section{Task Taxonomy}
\label{sec:task-taxonomy}

We describe each terminal-agent task in CLI-Universe along four orthogonal dimensions, distilled from patterns we observe across TB-Agent tasks: \emph{where the task lives} (Domain), \emph{what specialized knowledge the agent must apply} (Skill Types), \emph{what reasoning behaviors the task elicits} (Capabilities), and \emph{what engineering activity the agent is asked to perform} (Engineering Pillars). CLI-Universe uses this taxonomy to seed the random task-candidate assignment in its synthesis pipeline (Section~\ref{sec:blueprint_construction}), so that two tasks in the same domain still differ in the underlying capability pattern they exercise.

% -------------------------------------------------------------------
\subsection{Domain}
\label{sec:axis-domain}

The application area a terminal-agent task belongs to. Domains are human-curated and act as the entry point of the synthesis pipeline; each domain carries its own pool of allowed values on the other three dimensions, so the random combinations drawn from them remain realistic within that domain.

\begin{small}
\setlength{\tabcolsep}{4pt}
\renewcommand{\arraystretch}{1.25}
\begin{longtable}{@{}p{0.18\linewidth}>{\raggedright\arraybackslash}p{0.76\linewidth}@{}}
\caption{Domains in CLI-Universe.}
\label{tab:axis-domain}\\
\toprule
\textbf{Domain} & \textbf{Description} \\
\midrule
\endfirsthead
\multicolumn{2}{l}{\textit{Table~\thetable\ -- continued.}} \\
\toprule
\textbf{Domain} & \textbf{Description} \\
\midrule
\endhead
\bottomrule
\endlastfoot
Software Engineering   & General-purpose programming tasks: writing, modifying, or debugging application code in mainstream languages, including standalone scripts, libraries, and small services. \\
Debugging              & Localizing and repairing defects in existing artifacts---crashes, broken builds, corrupted state---where the agent must read failure signals to produce a targeted fix rather than write new functionality. \\
System Administration  & OS-level management tasks: process and service control, filesystem and permission management, scheduled jobs, and inspecting or repairing a running system. \\
% DevOps                 & Build, deploy, and runtime-infrastructure work: CI/CD pipelines, container and image management, environment provisioning, and release orchestration. \\
File Operations        & Manipulating, recovering, or transforming files and on-disk artifacts: extracting structured content from binary formats, large-scale text edits, or repairing corrupted containers, where filesystem- and format-level reasoning dominate. \\
Security               & Penetration-testing, capture-the-flag-style exploitation, secure-configuration audits, and vulnerability analysis where the agent must reason about attacker and defender behavior. \\
Data Processing        & ETL pipelines, log and text mining, schema reconciliation, and data wrangling across heterogeneous structured or semi-structured sources. \\
Data Querying          & Authoring queries against structured data backends such as SQL, SPARQL, or other declarative interfaces, where the difficulty lies in expressing intent in the query language rather than in transforming data downstream. \\
Data Science           & End-to-end analysis pipelines on real datasets: model evaluation harnesses, statistical inference, embedding and retrieval workflows, and reproducible analyses that combine data preparation with downstream interpretation. \\
Scientific Computing   & Numerical simulation, scientific data analysis, and computation-heavy workflows where correctness depends on physical or mathematical modeling rather than software engineering alone. \\
Mathematics            & Derivation- or proof-heavy tasks such as cryptanalysis, numerical linear algebra, and model-theoretic reasoning, where progress depends on mathematical insight before any code is written. \\
Optimization           & Formulating and solving constrained optimization problems---portfolio allocation, scheduling, parameter tuning---where the agent must select a solver and encode constraints faithfully. \\
Machine Learning       & Training, evaluation, and inference of models, including data preparation, hyperparameter sweeps, and small-scale serving setups. \\
Model Training         & Practical training and fine-tuning workflows: launching runs, recovering interrupted checkpoints, and scripting CLI training entry points, where the work centers on the training loop and its operational concerns. \\
Video Processing       & Frame-level video analysis and transformation, including extracting structured information from video streams and producing synchronized outputs across visual and audio modalities. \\
Web \& API Tooling     & Building or interacting with web services and browser-side flows: HTTP APIs, scraping, headless-browser automation, and integration with external services. \\
Games                  & Game-state reasoning and search-driven decision-making, where the agent implements or invokes engines that explore move spaces under explicit rules. \\
Personal Assistant     & User-facing tasks such as scheduling, planning under preferences and constraints, and producing concise actionable outputs from messy inputs. \\
\end{longtable}
\end{small}

% -------------------------------------------------------------------
\subsection{Skill Types}
\label{sec:axis-skills}

The specialized technical knowledge a terminal-agent task requires the agent to apply. Each task is tagged with the single load-bearing skill type; auxiliary skills (e.g.\ light shell glue around an algorithmic core) are not counted.

\begin{small}
\setlength{\tabcolsep}{4pt}
\renewcommand{\arraystretch}{1.25}
\begin{longtable}{@{}p{0.18\linewidth}>{\raggedright\arraybackslash}p{0.76\linewidth}@{}}
\caption{Skill Types, distilled from patterns observed across TB-Agent tasks.}
\label{tab:axis-skills}\\
\toprule
\textbf{Skill Type} & \textbf{Description} \\
\midrule
\endfirsthead
\multicolumn{2}{l}{\textit{Table~\thetable\ -- continued.}} \\
\toprule
\textbf{Skill Type} & \textbf{Description} \\
\midrule
\endhead
\bottomrule
\endlastfoot
Algorithmic      & Non-trivial algorithm or data-structure design where the core difficulty is the algorithm itself --- choosing, adapting, or combining primitives such as graph traversal, dynamic programming, custom indexing, or careful complexity reasoning. Routine glue around standard library calls does not qualify. \\
Data Processing  & Parsing, transforming, joining, or aggregating structured or semi-structured data (CSV/JSON/logs, tabular pipelines). The difficulty lies in handling schema variations, malformed records, or composing standard data operations correctly, rather than in inventing a new algorithm. \\
Systems          & Low-level OS, process, filesystem, or performance work whose correctness depends on system semantics --- signals, file descriptors, scheduling, memory layout, or precise concurrency behavior. Tagged when the agent must reason about how the OS or runtime actually behaves, not when it merely calls a high-level API that happens to live near the system. \\
Configuration    & Authoring or repairing configuration for tools, services, or build systems (Dockerfiles, systemd units, CI manifests, package or compiler config). The difficulty is in knowing the right knob and how knobs interact, not in writing program logic. \\
Shell Scripting  & End-to-end automation expressed primarily in shell, including control flow, pipelines, text-stream tools (\texttt{awk}, \texttt{sed}, \texttt{jq}), and quoting/escaping correctness. Tasks that only use shell to invoke a single Python or Node script do not qualify. \\
Mathematical     & Numerical or symbolic computation whose correctness rests on mathematical reasoning --- probability, linear algebra, signal processing, optimization, or numerical stability. Tagged when the agent must derive or pick the correct formulation, not when it merely plugs numbers into a known formula. \\
Deployment       & Packaging, building, releasing, or wiring up runtime infrastructure so that an artifact can actually run end-to-end: image building, dependency pinning, service registration, environment bootstrapping, or release orchestration. \\
Cryptography     & Cryptographic or cryptanalytic reasoning, including key recovery, protocol analysis, primitive misuse, and side-channel reasoning. Routine TLS or SSH configuration without analytic content is treated as Configuration, not Cryptography. \\
\end{longtable}
\end{small}

% -------------------------------------------------------------------
\subsection{Capabilities}
\label{sec:axis-cog}

The reasoning behaviors a terminal-agent task is intended to elicit during the agent's interaction with the environment. Each capability follows a strict \textsc{trigger / signature / not} rule applied to the trajectory, and a task may carry multiple capabilities.

\begin{small}
\setlength{\tabcolsep}{4pt}
\renewcommand{\arraystretch}{1.25}
\begin{longtable}{@{}p{0.18\linewidth}>{\raggedright\arraybackslash}p{0.76\linewidth}@{}}
\caption{Capabilities, distilled from patterns observed across TB-Agent tasks.}
\label{tab:axis-cog}\\
\toprule
\textbf{Capability} & \textbf{Description} \\
\midrule
\endfirsthead
\multicolumn{2}{l}{\textit{Table~\thetable\ -- continued.}} \\
\toprule
\textbf{Capability} & \textbf{Description} \\
\midrule
\endhead
\bottomrule
\endlastfoot
Exploration             & Probes an unfamiliar workspace (\texttt{ls}, \texttt{cat}, \texttt{find}, \texttt{grep}) before committing to a plan, because the task does not hand the agent a complete map of the environment and key affordances must be discovered from the filesystem itself. \\
Decomposition           & Enumerates sub-goals up front, before any irreversible action, so that subsequent steps execute against an explicit plan. Tagged when the trajectory shows a planning utterance with discrete sub-goals, not just a vague intent. \\
Error Recovery          & Reads an explicit failure signal (traceback, exit code, test output, log line) and revises the plan accordingly, rather than retrying the same action or escalating without diagnosis. \\
Specification Adherence & Returns to the original spec late in the trajectory to verify that the produced artifact matches the required format --- exact file path, output schema, argument signature --- catching cases where intermediate code drifts from the literal contract. \\
Working Memory          & Carries $\geq 5$ mutually dependent facts simultaneously across $10\!+\!$ turns without losing them, e.g.\ tracking variable bindings, file contents, and prior tool outputs that all need to remain consistent for the next step to succeed. \\
Long-Horizon Planning   & Sustains a coherent plan with inter-step dependencies over more than $10$ meaningful turns, where premature commitment to an early step closes off later ones. \\
Constraint Satisfaction & Maintains $\geq 2$ hard rules that pull in opposite directions, so progress on one risks breaking another --- e.g.\ a strict latency budget alongside a memory ceiling, or a format requirement that forbids the very tool the agent would otherwise reach for. \\
Modality Translation    & Translates information across semantic modalities, such as image$\to$code (rendering a diagram into a script), binary$\to$source (recovering structure from a stripped artifact), or table$\to$schema (inferring types and constraints from instance data). \\
Reverse Engineering     & Reconstructs hidden structure or intent from an opaque artifact, e.g.\ figuring out the calling convention of an undocumented binary, or inferring a service protocol from packet captures. \\
Mathematical Derivation & Produces a closed-form derivation or proof before any code is written, so the resulting implementation reflects a pre-justified formula rather than an empirically tuned approximation. \\
Adversarial Reasoning   & Reasons \emph{against} the designer's intent --- bypassing intended controls, finding exploitable corner cases, or mapping the attack surface of a system the agent is supposed to defend or analyze. \\
\end{longtable}
\end{small}

% -------------------------------------------------------------------
\subsection{Engineering Pillars}
\label{sec:axis-form}

The engineering pillar a terminal-agent task falls under --- whether building something new, fixing something broken, or restructuring something existing. Each task is tagged with a single dominant pillar.

\begin{small}
\setlength{\tabcolsep}{4pt}
\renewcommand{\arraystretch}{1.25}
\begin{longtable}{@{}p{0.18\linewidth}>{\raggedright\arraybackslash}p{0.76\linewidth}@{}}
\caption{Engineering Pillars, distilled from patterns observed across TB-Agent tasks.}
\label{tab:axis-form}\\
\toprule
\textbf{Pillar} & \textbf{Description} \\
\midrule
\endfirsthead
\multicolumn{2}{l}{\textit{Table~\thetable\ -- continued.}} \\
\toprule
\textbf{Pillar} & \textbf{Description} \\
\midrule
\endhead
\bottomrule
\endlastfoot
New feature creation    & Build a new artifact --- script, module, or service --- from scratch against a spec. The agent's primary work is to produce code that did not previously exist, rather than to modify or audit existing code. \\
Debugging               & Localize a defect from a failure signal and produce a minimal patch (error $\to$ root-cause $\to$ fix). The deliverable is a small, targeted change to existing code, not a rewrite. \\
Systems programming     & Low-level, OS-near, or performance-critical work where the deliverable lives close to the system layer --- a syscall wrapper, a kernel-adjacent utility, or a hand-tuned hot path --- and where systems-level concerns dominate the design rather than appear incidentally. \\
DevOps                  & Build, deploy, configure, or wire up runtime infrastructure end-to-end so that a target service or pipeline can actually run, including container/image work, CI/CD glue, and environment provisioning. \\
Feature iteration       & Extend or modify an existing codebase with a scoped new behavior, leaving the surrounding architecture intact. The change is additive or local rather than structural. \\
Large-scale refactoring & Cross-file structural change that preserves behavior --- e.g.\ extracting an interface across many modules, renaming a pervasive concept, or reorganizing a directory layout while keeping all tests green. \\
\end{longtable}
\end{small}

\section{Training Details}
\label{sec:training_details}

We fine-tune Qwen3 models using multi-turn SFT. Hyperparameters and hardware setup are summarized in Table~\ref{tab:training_hyperparameters}. Unless otherwise noted, all trajectories (including failed and incomplete ones) are retained; the effect of trajectory selection is studied in Section~\ref{sec:ablation_selection_subsec}.

\begin{table}[!h]
\centering
\caption{Training hyperparameters and hardware setup for SFT of Qwen3 students.}
\label{tab:training_hyperparameters}
\footnotesize
\setlength{\tabcolsep}{10pt}
\renewcommand{\arraystretch}{1.1}
\begin{tabular}{ll}
\toprule
\textbf{Hyperparameter} & \textbf{Value} \\
\midrule
Optimizer & AdamW ($\beta_2 = 0.95$) \\
Peak learning rate & $1 \times 10^{-5}$ \\
LR scheduler & cosine\_with\_min\_lr (min $1 \times 10^{-6}$) \\
Warmup ratio & 0.03 \\
Weight decay & 0.01 \\
Max gradient norm & 1.0 \\
Epochs & 5 \\
Per-device batch size & 1 \\
Gradient accumulation steps & 4 \\
Context length & 64K \\
Precision & bf16 \\
Sequence parallelism & degree 2 \\
GPUs & 32 $\times$ NVIDIA H200 \\
Global batch size & 64 \\
\bottomrule
\end{tabular}
\end{table}

\section{Failure Mode Examples}
\label{sec:failure-mode-examples}

\newcolumntype{Y}{>{\raggedright\arraybackslash}X}

Each failure mode is illustrated with one CLI-Universe-32B trajectory on Terminal-Bench~2, shown as a turn-level timeline. A ``turn'' is one agent action cycle (reasoning $\to$ command $\to$ observation).

% -------------------------------------------------------------------
\subsection{Disobey Task Specification (Execution)}
\label{sec:fm-disobey}

\textbf{Task (sam-cell-seg):} Write \texttt{convert\_masks.py}. Task specifies: \texttt{-{}-output\_path /app/test\_output.csv} (a \texttt{.csv} file path). 21 turns.

{\small
\setlength{\tabcolsep}{3pt}
\renewcommand{\arraystretch}{1.15}
\begin{tabularx}{\linewidth}{lYY}
\toprule
\textbf{Turn} & \textbf{Agent Action} & \textbf{Failure Relevance} \\
\midrule
6 & Drafts: \texttt{os.path.dirname(} \newline \texttt{args.output\_path)} + \newline \texttt{result.to\_csv(args.output\_path)} & Correctly treats \texttt{output\_path} as \textbf{file} \\
12 & Rewrites: \texttt{os.makedirs(} \newline \texttt{args.output\_path)} + \newline \texttt{os.path.join(args.output\_path, ...)} & \ding{55} Switches to \textbf{directory} handling \\
12 & Tests with \newline \texttt{-{}-output\_path /app/output} & \ding{55} Tests with directory, not task's \texttt{/app/test\_output.csv} \\
13--16 & Same \texttt{os.makedirs} + \texttt{os.path.join} \newline across 4 rewrites & \ding{55} Directory convention persists \\
13--16 & Each rewrite tested with \newline \texttt{-{}-output\_path /app/output} & \ding{55} Never tests with the task's file path \\
20 & \texttt{task\_complete: true}, \newline ``The task is fully complete'' & Spec violation never exposed \\
\bottomrule
\end{tabularx}
}

\smallskip\noindent Early draft (Turn~6) had correct file-path handling. From Turn~12 onward, all landed versions use directory semantics, and all 4 test invocations used \texttt{/app/output}, and the task's \texttt{.csv} file path was never tested.

% -------------------------------------------------------------------
\subsection{Step Repetition (Execution)}
\label{sec:fm-repetition}

\textbf{Task (build-pov-ray):} Download and compile POV-Ray 2.2 from source. 357 turns.

{\small
\setlength{\tabcolsep}{3pt}
\renewcommand{\arraystretch}{1.15}
\begin{tabularx}{\linewidth}{lYY}
\toprule
\textbf{Turn} & \textbf{Agent Action} & \textbf{Failure Relevance} \\
\midrule
$\sim$1--30 & \texttt{curl -s 'http://.../download/'} \newline \texttt{| grep -o 'http[\^{}"]+'  | head -20} & No download link found \\
$\sim$100 & Same \texttt{curl} command & \ding{55} Same empty result, $\sim$100th attempt \\
$\sim$150 & Agent reasons: ``switch to archive.org'' & Strategy change in \textbf{reasoning only} \\
$\sim$151 & \texttt{curl -s 'http://.../download/' ...} & \ding{55} Same command despite reasoning \\
$\sim$1--357 & \texttt{find / -maxdepth 5 -type f} \newline \texttt{| grep -i 'pov'} $\times$151 & \ding{55} Same filesystem state each time \\
$\sim$1--357 & \texttt{grep -rl '2\textbackslash.2'} \newline \texttt{/app/povray-source/} $\times$44 & \ding{55} Same files listed each time \\
Final & \texttt{curl} $\times$165, \texttt{find} $\times$151, \texttt{grep} $\times$44 & Same commands loop without convergence \\
\bottomrule
\end{tabularx}
}

\smallskip\noindent Reasoning mentions strategy changes, but actual commands never change. The same \texttt{curl}/\texttt{find}/\texttt{grep} loop runs 44--165 times each across 357 turns.

% -------------------------------------------------------------------
\subsection{Unaware of Termination (Execution)}
\label{sec:fm-termination}

\textbf{Task (compile-compcert):} Compile CompCert. Environment has hard 120s timeout (exit 124). 159 turns, 0 successes.

{\small
\setlength{\tabcolsep}{3pt}
\renewcommand{\arraystretch}{1.15}
\begin{tabularx}{\linewidth}{lYY}
\toprule
\textbf{Turn} & \textbf{Agent Action} & \textbf{Failure Relevance} \\
\midrule
$\sim$1 & \texttt{opam init -{}-compiler=4.05.0} \newline $\to$ exit 124 & First timeout (hard 120s ceiling) \\
$\sim$5 & Requests duration 60s $\to$ exit 124 & \ding{55} Ceiling unchanged \\
$\sim$10 & Requests duration 120s $\to$ exit 124 & \ding{55} Still 120s ceiling \\
12 & \texttt{echo hello} (duration 2s) \newline $\to$ exit 124 & Even trivial commands time out (systemic issue) \\
$\sim$30 & Requests duration 300s $\to$ exit 124 & \ding{55} Escalation begins \\
$\sim$60 & Requests duration 1200s $\to$ exit 124 & \ding{55} 10$\times$ original, same result \\
$\sim$80 & Agent: ``environment is extremely slow'' \newline $\to$ duration 3600s & \ding{55} Wrong diagnosis \\
$\sim$100 & Writes one-shot \texttt{/tmp/build.py} \newline $\to$ exit 124 & \ding{55} Condensing into one script doesn't help \\
$\sim$120 & Duration 36000s (10 hours) \newline $\to$ exit 124 & \ding{55} \\
$\sim$146 & Duration 72000s (20 hours) \newline $\to$ exit 124 & \ding{55} Final attempt, 600$\times$ original timeout \\
\bottomrule
\end{tabularx}
}

\smallskip\noindent 159 turns, 163 commands, 0 successes. Agent escalates duration from 1s to 72,000s, but the hard 120s ceiling never changes. Even \texttt{echo hello} fails.

% -------------------------------------------------------------------
\subsection{Context Loss (Coherence)}
\label{sec:fm-context-loss}

\textbf{Task (make-mips-interpreter):} Build a MIPS interpreter in JavaScript to run a DOOM binary. 300 turns.

{\small
\setlength{\tabcolsep}{3pt}
\renewcommand{\arraystretch}{1.15}
\begin{tabularx}{\linewidth}{lYY}
\toprule
\textbf{Turn} & \textbf{Agent Action} & \textbf{Failure Relevance} \\
\midrule
5--6 & \texttt{readelf -h/l/s} on \newline \texttt{/app/doomgeneric\_mips} $\to$ \newline ``I have the function addresses \newline and entry point'' & Facts first read \\
10--11 & Re-runs \texttt{readelf -s | grep '\_\_start'} \newline $\to$ \texttt{0x400110} & Had to re-check: didn't retain \newline numeric value from T6 \\
13 & Summarizes: \texttt{\_\_start} at \texttt{0x400110}, \newline BSS \texttt{0x004a6ba0}--\texttt{0x004b0a40} & Facts stated explicitly \\
22 & Hardcodes \texttt{TEXT\_START}, \texttt{DATA\_START}, \newline \texttt{BSS\_START}, \texttt{STACK\_START} in \texttt{vm.js} & Constants committed to file \\
49--64 & Re-runs readelf, re-checks entry point, \newline re-states \texttt{0x400110} + BSS layout & \ding{55} Re-checking facts already \newline established at T11--13 \\
96--103 & Re-reads \texttt{f.seek(0x400110)} bytes, \newline re-runs readelf, rewrites \texttt{vm.js} \newline with same constants & \ding{55} Same facts re-derived again \\
131 & ``I defined \texttt{read8} but forgot to \newline define \texttt{write8}'' $\to$ patches with \texttt{sed -i} & Bug diagnosed, fix attempted \\
145 & ``I defined \texttt{read8} but forgot to \newline define \texttt{write8}'' again & \ding{55} Same diagnosis, same \texttt{sed} patch, \newline 14 turns later \\
170--178 & Three consecutive turns re-examine \newline \texttt{0x439ad0} / \texttt{0xa0450000}, re-argue \newline endianness, re-derive big-endian reads & \ding{55} Same evidence, same conclusion, \newline 3 times \\
298--300 & Summarizes ELF structure and entry point \newline $\to$ rewrites \texttt{vm.js} with same \newline \texttt{e\_phoff}/\texttt{e\_entry} parser & \ding{55} Same parser that should have \newline been the base throughout \\
\bottomrule
\end{tabularx}
}

\smallskip\noindent The agent repeatedly loses established facts across turns: entry point \texttt{0x400110} is re-checked via readelf at least 4 times, \texttt{write8} bug is diagnosed and patched twice, endianness is re-derived 3 times. Despite hardcoding constants in \texttt{vm.js} at Turn~22, the agent keeps re-running the same diagnostic commands to re-establish the same values.

% -------------------------------------------------------------------
\subsection{Task Derailment (Coherence)}
\label{sec:fm-derailment}

\textbf{Task (regex-chess):} Write \texttt{/app/re.json}, regex pairs for chess move generation via \texttt{re.sub}. 10 turns.

{\small
\setlength{\tabcolsep}{3pt}
\renewcommand{\arraystretch}{1.15}
\begin{tabularx}{\linewidth}{lYY}
\toprule
\textbf{Turn} & \textbf{Agent Action} & \textbf{Failure Relevance} \\
\midrule
1 & \texttt{cat /app/check.py}, understands \newline need for \texttt{[regex, replacement]} pairs & Reads requirement \\
2--3 & \texttt{python3 -c "import chess;} \newline \texttt{for move in board.legal\_moves:} \newline \texttt{print(move.uci())"} & \ding{55} Diagnostic, enumerates moves \\
4--5 & \texttt{python3 -c "board.push(move);} \newline \texttt{print(board.fen())"} & \ding{55} Diagnostic, FEN after each move \\
6--8 & Manual FEN parsing, confusion \newline about rank ordering & \ding{55} Diagnostic, notation study \\
9--10 & More \texttt{python3} probes on \newline castling, en passant & \ding{55} \texttt{/app/re.json} \textbf{never written} \\
\bottomrule
\end{tabularx}
}

\smallskip\noindent 10 turns of diagnostic \texttt{python-chess} scripts. 0 turns of implementation. The deliverable is never started.

% -------------------------------------------------------------------
\subsection{Reasoning--Action Mismatch (Coherence)}
\label{sec:fm-mismatch}

\textbf{Task (feal-differential-cryptanalysis):} Implement differential attack on 4-round FEAL. 35 turns.

{\small
\setlength{\tabcolsep}{3pt}
\renewcommand{\arraystretch}{1.15}
\begin{tabularx}{\linewidth}{lYY}
\toprule
\textbf{Turn} & \textbf{Agent Action} & \textbf{Failure Relevance} \\
\midrule
2 & Reads \texttt{/app/feal.py}; reasoning contains \newline \texttt{getright()} + \texttt{seed*1234567} brute-force & Forward brute-force present \newline from the start \\
3 & Reasons: ``work \textbf{backwards} through \newline round~4'', but embedded code still uses \newline \texttt{getright()} + brute-force & \ding{55} Reasoning says backward, \newline code template is forward \\
4 & Discovers differential: \newline ``\texttt{0x80800000} $\to$ \texttt{0x02000000}'' & Correct differential found \\
5 & Reasons: ``the key differential \newline property is...'', code still \newline \texttt{getright()} + brute-force & \ding{55} Backward reasoning, \newline forward code \\
7 & Writes \texttt{attack.py}, brute-force \newline over large key space, too slow & \ding{55} Still forward \\
10 & Rewrites \texttt{attack.py}, still brute-force & \ding{55} Still forward \\
11 & Reasons: ``My earlier analysis was wrong. \newline Let me re-examine'' + backward analysis & Retracts, starts over \\
14 & Writes \texttt{attack.py}, forward search again & \ding{55} Still forward \\
19 & Reasons: ``My earlier analysis was wrong'' \newline again + backward analysis & Second retraction \\
22 & Writes \texttt{attack.py}, forward search again & \ding{55} Still forward \\
31 & Reasons: ``Now I understand the encryption \newline structure'' + backward analysis & Third attempt at backward plan \\
35 & Final \texttt{attack.py}: \newline \texttt{for seed in range(65536):} \newline \texttt{key5 = (seed*1234567)} + \newline \texttt{f\_function(getright(plain[i]) \^{} key5)} & \ding{55} \textbf{Still forward brute-force} \\
\bottomrule
\end{tabularx}
}

\smallskip\noindent The mismatch is present from Turn~2: the agent's reasoning text describes backward differential propagation, but the code template embedded in the same reasoning uses \texttt{getright()} + \texttt{seed*1234567} forward brute-force. Across 35 turns and 3 retraction cycles, this contradiction never resolves.

% -------------------------------------------------------------------
\subsection{Premature Termination (Verification)}
\label{sec:fm-premature}

\textbf{Task (cobol-modernization):} Modernize a COBOL program to Python, producing identical \texttt{.DAT} output files. 20 turns.

{\small
\setlength{\tabcolsep}{3pt}
\renewcommand{\arraystretch}{1.15}
\begin{tabularx}{\linewidth}{lYY}
\toprule
\textbf{Turn} & \textbf{Agent Action} & \textbf{Failure Relevance} \\
\midrule
1--8 & Reads COBOL source, analyzes \newline record layouts & Understanding the program \\
9 & Writes \texttt{/app/program.py}, first version & Book owner update uses \newline \texttt{offset:offset+4} (bytes 0--3, \newline the \textbf{book-ID} field) \\
9 & Should write to \texttt{offset+24:offset+28} \newline (the \textbf{owner} field) & \ding{55} Wrong offset: overwrites \newline book-ID instead of owner \\
11--16 & Rewrites \texttt{program.py} 4 more times, \newline book owner update still uses \newline \texttt{offset:offset+4} & \ding{55} Bug persists through all rewrites \\
17 & Runs comparison: \newline \texttt{print('Books equal:', b1==b2)}, \newline but \texttt{b1} undefined (NameError) & \ding{55} Verification script has variable \newline scope bug, produces no valid result \\
17 & Agent notes: ``typo in variable names \newline (\texttt{a1}, \texttt{b1}, \texttt{t1} were not defined)'' & Recognizes NameError but doesn't \newline fix and re-run \\
19--20 & \texttt{task\_complete: true}, declares success & \ding{55} No valid verification; book-ID \newline still corrupted in output \\
\bottomrule
\end{tabularx}
}

\smallskip\noindent The book-owner update writes to the wrong byte offset (0--3 instead of 24--27) across \textbf{all 5 versions} of the code. The verification script has a NameError and produces no result. The agent acknowledges the NameError but declares completion without re-running a corrected verification, so the core bug (wrong offset) is never caught.

% -------------------------------------------------------------------
\subsection{No or Incorrect Verification (Verification)}
\label{sec:fm-incorrect-verif}

\textbf{Task (cancel-async-tasks):} Async task queue with correct SIGINT/KeyboardInterrupt cleanup. 6 turns.

{\small
\setlength{\tabcolsep}{3pt}
\renewcommand{\arraystretch}{1.15}
\begin{tabularx}{\linewidth}{lYY}
\toprule
\textbf{Turn} & \textbf{Agent Action} & \textbf{Failure Relevance} \\
\midrule
1 & Writes \texttt{/app/run.py}, on SIGINT: \newline queue \texttt{None} sentinels, await workers & Implementation has subtle \newline race condition \\
2 & Test v1: \newline \texttt{asyncio.create\_task(} \newline \texttt{run\_with\_interrupt())} & \ding{55} Structurally wrong, \newline doesn't simulate SIGINT \\
3 & Test v2: \newline \texttt{asyncio.wait\_for(} \newline \texttt{run\_with\_interrupt(), timeout=0.5)} & \ding{55} Raises \textbf{TimeoutError}, \newline not \textbf{KeyboardInterrupt} \\
4 & Test v2 passes, ``all 5 cleanups ran'' & \ding{55} \textbf{TimeoutError} path verified, \newline not \textbf{SIGINT} path \\
5 & ``Cleanup verified via timeout \newline simulation'' $\to$ \texttt{task\_complete: true} & \ding{55} Claims SIGINT verified \\
6 & Reaffirms completion & Real bug: sentinels queued \newline behind items $\to$ never exposed \\
\bottomrule
\end{tabularx}
}

\smallskip\noindent 3 test iterations, each testing the wrong exception path. \texttt{asyncio.wait\_for} raises \texttt{TimeoutError}, not \texttt{KeyboardInterrupt}, so the real SIGINT race condition is never triggered.

% -------------------------------------------------------------------
\subsection{Weak Verification (Verification)}
\label{sec:fm-weak-verif}

\textbf{Task (build-cython-ext):} Build pyknotid 0.5.3 Cython extensions, numpy pinned to 2.3.0. Evaluator checks 11 properties. 26 turns.

{\small
\setlength{\tabcolsep}{3pt}
\renewcommand{\arraystretch}{1.15}
\begin{tabularx}{\linewidth}{lYY}
\toprule
\textbf{Turn} & \textbf{Agent Action} & \textbf{Failure Relevance} \\
\midrule
1--2 & Clones repo, checks numpy $\to$ 2.3.0 & \checkmark numpy version checked \\
3 & \texttt{build\_ext -{}-inplace}, \newline \texttt{cinvariants} hangs/times out & $\triangle$ Partial build \\
4--9 & Fixes \texttt{np.float} $\to$ \texttt{float}, \newline \texttt{np.complex} $\to$ \texttt{complex}, \newline \texttt{gcd} import & Numpy 2.0 compat \\
10 & README snippet test \newline $\to$ returns \texttt{6.999999999999998} & \checkmark 2 functions checked \\
11--12 & \texttt{pytest tests/} $\to$ 18 passed & \checkmark Test suite passes \\
13 & \texttt{pip install -e /app/pyknotid} \newline \texttt{-{}-force-reinstall} & $\triangle$ May overwrite numpy 2.3.0 \\
14--25 & Tries \texttt{.so} file check + individual \newline Cython function tests & \ding{55} Malformed commands, \newline never execute \\
26 & \texttt{task\_complete: true} & \ding{55} No re-check of numpy; \newline \texttt{cinvariants} never verified \\
\bottomrule
\end{tabularx}
}

\smallskip\noindent Verified 4/11 properties. \texttt{-{}-force-reinstall} at Turn~13 invalidates the numpy check from Turn~1. \texttt{cinvariants} timed out during build and was never re-verified. Last verification commands had JSON escaping errors and never actually ran.

\end{document}